\documentclass{llncs}
\usepackage[dvips]{graphics}
\usepackage{epsfig}
\usepackage{amsfonts}
\usepackage{amsmath}
\usepackage{color}
\usepackage{url}
\usepackage{multirow}
\usepackage{subfigure}
\usepackage{psfrag}

\usepackage{algorithmic,algorithm}

\begin{document}

\pagestyle{empty}

\mainmatter

 \title{Local Optima Networks, Landscape Autocorrelation  and \\ Heuristic Search Performance}

\newif\ifshowchanges
\newcommand{\change}[1]{\ifshowchanges\textcolor{red}{#1}\else#1\fi}

\newif\ifshowcomments
\newcommand{\comment}[1]{\ifshowcomments\textbf{(#1)}\fi}

\author{ Francisco Chicano\inst{1} \and Fabio Daolio\inst{2} \and  Gabriela Ochoa\inst{3} \and S\'ebastien V\'erel\inst{4} \and Marco Tomassini\inst{2}\and Enrique Alba\inst{1}}

\authorrunning{F. Chicano et al.}   

\institute{E.T.S. Ingenier\'ia Inform\'atica, University of M\'alaga, Spain. \and Information Systems Department, University of Lausanne, Lausanne, Switzerland. \and School of Computer Science, University of Nottingham, Nottingham, UK. \and INRIA Lille - Nord Europe and University of Nice Sophia-Antipolis, France.}

\maketitle

\begin{abstract}
Recent developments in fitness landscape analysis include the study of Local Optima Networks (LON) and applications of  the Elementary Landscapes theory. This paper represents a first step at combining these two tools to explore their ability to forecast the performance of search algorithms. We base our analysis on the Quadratic Assignment Problem (QAP) and conduct a large statistical study over 600  generated instances of different types. Our results reveal  interesting links  between the network measures, the autocorrelation measures and the performance of heuristic search algorithms.
\end{abstract}


\section{Introduction}

An improved understanding of  the structure of  combinatorial fitness landscapes can facilitate the design and further successful application of heuristic search methods to solve hard computational problems. This article brings together two recent developments in fitness landscape analysis for combinatorial optimisation, namely, local optima networks (LONs) and elementary landscape decomposition. LONs represent a new model of combinatorial landscapes based on the idea of compressing the information given by the whole problem configuration space into a smaller mathematical object that is the graph having as vertices the local optima  and as edges the possible transitions between them \cite{pre09,ieee11}. This characterization of landscapes as complex networks enables the use of tools and metrics of the complex networks domain \cite{barrat2008dynamical} and has brought new insights into the global structure of the landscapes studied in the past~\cite{QAPcec10}. 

The QAP has been recently analysed using this model \cite{QAPcec10} and the clustering structure of the local optima networks of  two classes of  QAP instances  was studied in \cite{PhysA10}. The study revealed that the so called ``real-like''  instances have significantly more optima cluster (or modular) structure than the class of random uniform instances of the QAP problem. Using the theory of elementary landscapes \cite{Barnes2002} the QAP has been analysed in \cite{chicano2010-qap} and the methodology  presented in  \cite{Chicano2011ecj} has been used to decompose the landscape and provide expressions for each elementary component. This decomposition can then be used to exactly compute the  autocorrelation coefficient and the autocorrelation length  of any arbitrary QAP instance \cite{Chicano2012aml}.

In this article, the expression in \cite{Chicano2012aml} is used to calculate the autocorrelation length of the two classes of QAP instances studied in \cite{PhysA10}. Since for those instances the LONs were exhaustively computed, the exact number of local optima are known in all cases. This will allow us to support the autocorrelation length conjecture~\cite{Stadler2002}, which links the autocorrelation length to the number of local optima of a landscape. We also conduct a correlation study among several network metrics calculated on the extracted LONs and the success rate of two heuristic search algorithms: simulated annealing and genetic algorithm. Our goal is to discover relationships between fitness landscape features and the performance of heuristic search methods.

The article is structured as follows. Section \ref{methods} includes the relevant definitions, methodologies and metrics used in this article.  Section \ref{study} presents the correlation study and Section \ref{discussion} discusses our main findings and suggests directions for future work.

\section{Background}
\label{methods}

In this section we introduce all the background concepts required in the rest of the paper. We define the QAP, describe the LONs, introduce the network metrics, the autocorrelation length and describe the heuristic search algorithms used in the experimental section.
 
\subsection{The Quadratic Assignment Problem} 
\label{qap}

The QAP is a combinatorial problem in which a set of facilities with given flows have to be assigned to a set of locations with given distances in such a way that the sum of the product of flows and distances is minimized. A solution to the QAP is generally written as a permutation $\pi$ of the set $\{1,2,...,n\}$. The cost associated with a permutation $\pi$ is: $C(\pi)=\sum_{i=1}^{n}\sum_{j=1}^{n}{a_{ij}b_{\pi_{i}\pi_{j}}}$,  where $n$ denotes the number of facilities/locations and  $A=(a_{ij})$ and $B=(b_{ij})$ are referred to as the distance and flow matrices, respectively. The contents of these two matrices  characterize the class of instances of the QAP problem.

For the statistical analysis conducted here,  the two instance generators proposed in~\cite{Knowles2003emo} for the multi-objective QAP were adapted for the single-objective QAP. The first generator produces uniformly random instances where all flows and distances are integers sampled from uniform distributions.  The second generator produces flow entries that are non-uniform random values. The instances produced have the so called ``real-like'' structure since they resemble the structure of QAP problems found in practical applications. We consider here these two types of instances and  tree problem dimensions:  9, 10 and 11.  Therefore, we have  six different instance groups. For each group, 100  instances were generated for a total of 600 QAP instances that will be used in our study.

\subsection{Local Optima Networks}

In order to define the local optima network of the QAP instances, we need to provide the definitions for the nodes and  edges of the network. The vertices of the graph can be straightforwardly defined as the local minima of the landscape. In this  work,  we select small QAP instances such that it is feasible to obtain the nodes  exhaustively by running a best-improvement local search algorithm from every configuration (permutation) of the search space. The neighborhood of a configuration is defined by the {\em pairwise exchange } or {\em swap} operation, which is the most basic operation used by many metaheuristics for QAP. This operator simply exchanges any two positions in a permutation, thus transforming it  into another permutation. The neighborhood size is thus $|V(s)| = n(n-1)/2$. Given a local optima $s$, its \emph{basin of attraction} is defined as the set of solutions $s'$ from which $s$ can be reached using a hill-climbing algorithm~\cite{QAPcec10}.

The edges account for the transition probability between basins of attraction of the local optima. More formally, the edges reflect the total probability of going from basin $b_i$ to basin $b_j$, which is the average over all $s \in b_i$ of the transition probabilities  to solutions $s' \in b_j$. The reader is referred to \cite{QAPcec10} for a more detailed exposition.

We define a \textit{Local Optima Network} (LON) as being the graph  $G=(S^*,E)$ where the set of vertices $S^*$ contains all the local optima, and there is an edge $e_{ij} \in E$ with weight $\vec{w}_{ij} = p(b_i \rightarrow b_j)$ between two nodes $i$ and $j$ if and only if $p(b_i \rightarrow b_j) > 0$, where $p(b_i \rightarrow b_j)$ is the probability of moving from basin $b_i$ to basin $b_j$ in one step. Notice that since each optimum has its associated basin, $G$ also describes the interconnection of basins.

\subsection{Network metrics}

We describe below the six network metrics considered  in our analysis.

\begin{description}

\item[Number of vertices, $N_v$ :]  The number of nodes of a LON is simply the number of local optima in the fitness landscape. It is exhaustively computed running a best-improvement hill-climbing algorithm from each solution of the search space.

\item [Clustering coefficient, $Cc$ :] Measures the probability that two neighbors of a given node are also neighbors of each other~\cite{barrat2008dynamical}. In other words, it accounts for the ratio of connected triples in the graph. In the language of social networks, it measures how likely it is that  the friend of your friend is  also your friend. 

\item [Shortest path length to the optimum, $L_{opt}$:] A standard metric to characterize the structure of networks is the shortest path length (number of link hobs) between two nodes in the network. In order to compute this measure on the LONs, we considered the expected number of moves (in the case of QAP swap moves)  to pass from one basin to the other. This expected number can be computed by considering the inverse of the transition probabilities between basins: $1/w_{ij}$. We use this to calculate the average shortest paths leading to the global optimum.

\item [Disparity, $Y_{2}$:]  Measures  the local heterogeneity introduced by edge weights~\cite{barrat2008dynamical}. It indicates whether the outgoing links  from a given  node have mostly the same weights (transition probabilities) or there is one outweighing the others. Disparity for a vertex $i$ is computed as $Y_2(i)=\sum_{j \neq i} (w_{ij}/s_i)^2$, where $s_i=\sum_{j\neq i} w_{ij}$ is the so-called \emph{strength} of vertex $i$.

\item [Fitness-fitness correlation, $F_{nn}$:]  Measures the correlation between the fitness values of adjacent local optima. More precisely, we estimate with the Spearman rank correlation coefficient the correlation between the fitness value $f_i$ of vertex $i$ and its weighted-average nearest-neighbors fitness, defined as $F_{nn}^{w}(i)=1/s_i \sum_{j \neq i} w_{ij} f_j$. 

\item [Modularity, $Q$:] Clusters or communities in networks can be loosely defined as groups of nodes that are strongly  connected between them and poorly connected with the rest of the graph. To calculate the level of community structure, also known as modularity, we consider a graph clustering algorithm that is based on the simulation of network flow~\cite{mcl}, as in~\cite{PhysA10}.

\end{description}

\vspace{-6pt}
\subsection{Calculation of the autocorrelation length}
\vspace{-6pt}

Let us consider an infinite random walk $\{x_0,x_1,\ldots \}$ on the
solution space such that $x_{i+1} \in N(x_i)$. The \emph{random walk
autocorrelation function} $r:\mathbb{N} \rightarrow \mathbb{R}$ is
defined as~\cite{Weinberger1990}:
\begin{equation}
r(s) = \frac{\left\langle f(x_t)f(x_{t+s})\right\rangle_{x_0,t}-\left\langle f(x_t) \right\rangle^2_{x_0,t}}
{\left\langle f(x_t)^2 \right\rangle_{x_0,t} - \left\langle f(x_t) \right\rangle_{x_0,t}^2}
\end{equation}
where the subindices $x_0$ and $t$ indicate that the averages are computed over all the starting solutions $x_0$
and along the complete random walk.
The \emph{autocorrelation length} $\ell$~\cite{GarciaPelayo:1997p2070} is defined as
$\ell = \sum_{s=0}^{\infty} r(s)$. Using the landscape decomposition of the QAP in \cite{Chicano2012aml} the authors provide a closed-form formula for $\ell$ based on the matrices $(a_{ij})$ and $(b_{ij})$ of the QAP instance. We will use in the present article this formula to efficiently compute the autocorrelation length of all the instances in our experimental study.

\vspace{-12pt}
\subsection{Heuristic search algorithms and the performance metric}

We considered two well-known heuristic search algorithms: simulated annealing (SA) and genetic algorithm (GA). The SA uses a cooling factor of   $0.9983$ and an  initial temperature of $10^7$. The neighborhood move is the same used for generating the LONs, namely, the pairwise exchange or swap operation in permutation space. The GA is a steady-state GA with a population size of 100, where one solution is computed at a time and inserted in the population using elitist replacement.  The individuals are selected using a binary tournament. The genetic operators are  the partially mapped crossover (PMX) \cite{Back2000} and the pairwise exchange mutation operation applied with  probability $0.3$. We perform 100 independent runs for each algorithm and instance.

In order to measure the performance of a search algorithm solving the QAP instances we use the success (hit) rate, defined as the fraction of the 100 independent runs that found the global optimum. Both algorithms stop when they reach $10\;000$ function evaluations.

\vspace{-12pt}
\section{Correlation Study}
\label{study}
\vspace{-12pt}

Our statistical analysis considers the pair-wise correlation among  the six network metrics, the autocorrelation length, and the SA and GA success rates. As mentioned above, six classes of instances are considered, including  two types  (`uniform' and `real-like')  of instances as described in Section \ref{qap}, and 3 problem sizes 9, 10 and 11. Each of the 6 instance classes  is considered separately, and 100 instances conform the sample for the statistical analysis in each class.

\begin{figure}[!ht]
\centering
\includegraphics[width=0.83\textwidth]{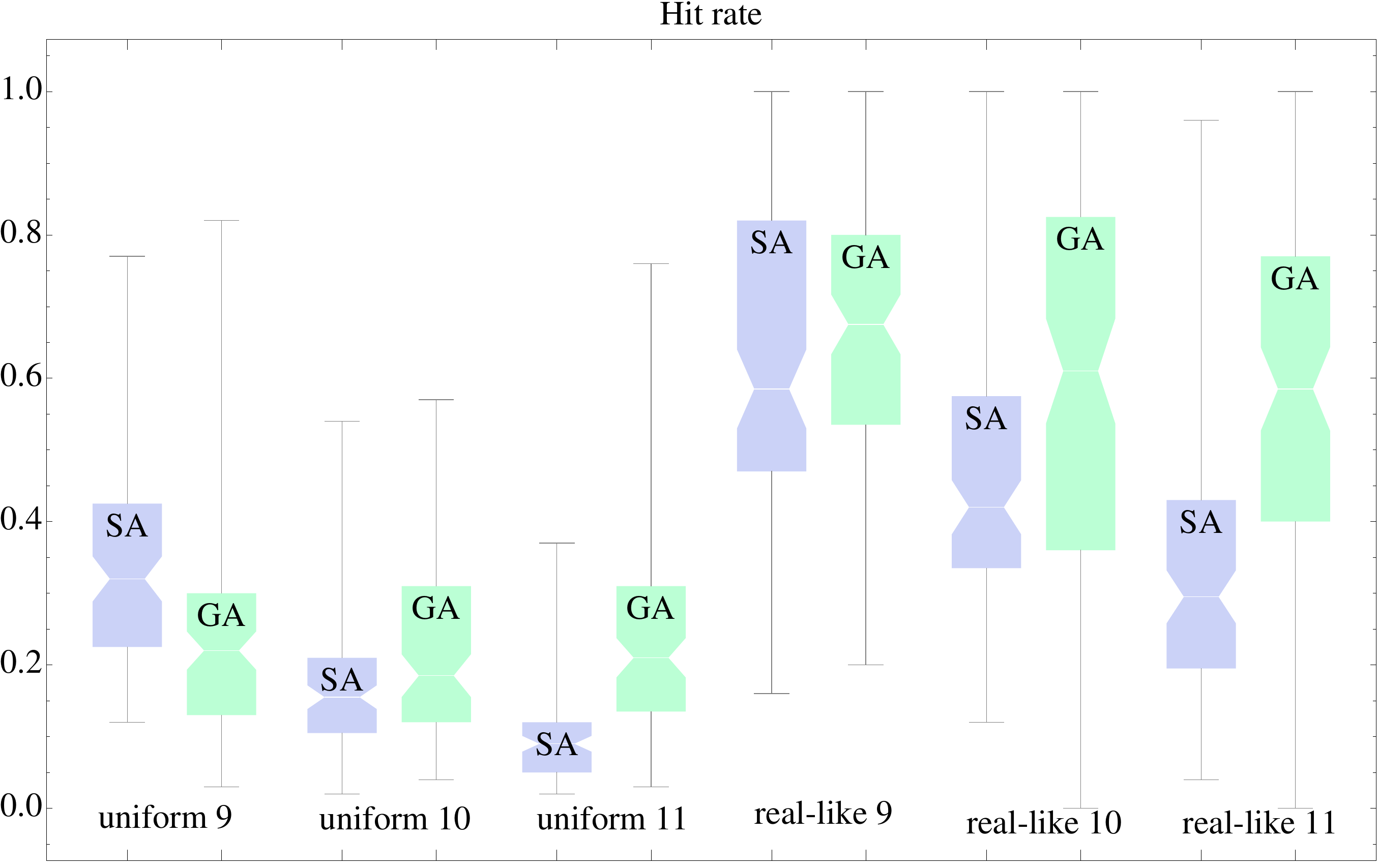}
\caption{Hit rate of the GA and SA for the 6 classes of instances considered. The boxplots are computed on the results of the algorithms over the 100 random instances per class.}
\label{fig:boxplots}
\vspace{-12pt}
\end{figure}

The main goal of our study is to discover whether some of the studied metrics can predict the performance of a heuristic search algorithm on a given instance class. We start, then, by showing the performance of the two selected search algorithms: SA and GA. Figure~\ref{fig:boxplots} illustrate the range and distribution of the hit rates for each algorithm and instance class, while Table~\ref{tab:hitrate} contains hit rate values, number of local optima and the average shortest path length for the instance classes. We show the values of these two network metrics because they seem to have an impact on the hit rate, as we will see later.

The results suggest that, for each instance type (uniform or real-like),  the hit rates for both algorithms decrease as the instance size increases, with the only exception of GA in the real-like 11 class. This is expected as the size of the search space increases, and so locating the global optimum is harder. However, the hit rates for real-like instances are much higher in all cases, which confirms that these instances are easier to solve for both algorithms~\cite{QAPcec10}. In Table~\ref{tab:hitrate} we can see that real-like instances have a lower number of local optima compared to uniform instances, which explains why real-like instances are easier to solve than the uniform ones. A second observation is that the hit rate of the GA, for a given instance type, does not change much when the size of the instances is increased. However, in SA the hit rate is clearly reduced when the size increases. That is, SA seems to be quite sensitive to the size of the instance (in addition to the type) while GA is clearly sensitive to the type of the instance (uniform or real-like) but little sensitive to the size.


\begin{table}[!ht]
\caption{Number of local optima ($N_v$), shortest path to the optimum ($L_{opt}$) and hit rate of the GA and SA for the 6 instance classes. We show the average and the standard deviation.}
\label{tab:hitrate}
\centering
\begin{tabular}{|l|l|r|r|r|r|r|r|r|r|}
\hline
\multicolumn{2}{|c}{} & \multicolumn{2}{|c|}{{$\mathbf{N_v}$}} & \multicolumn{2}{|c|}{{$\mathbf{L_{opt}}$}} & \multicolumn{2}{|c|}{\textbf{GA}} & \multicolumn{2}{|c|}{\textbf{SA}} \\
\cline{3-10}
\multicolumn{2}{|c|}{\textbf{Class}}	 & Avg. & Std. dev. & Avg. & Std. dev. & Avg. & Std. dev. & Avg. & Std. dev.\\
\hline
\multirow{3}{*}{uni} & $n=9$  & 131.220 & 51.268   & 25.761 & 11.231 & 0.220 & 0.170 & 0.320 & 0.200 \\
                     & $n=10$ & 399.840 & 153.097  & 45.217 & 17.120 & 0.185 & 0.190 & 0.155 & 0.105 \\
                     & $n=11$ & 1337.300 & 453.520 & 76.815 & 26.698 & 0.210 & 0.175 & 0.090 & 0.070 \\
\hline
\multirow{3}{*}{rl}  & $n=9$  & 14.300 & 7.473  & 8.564 & 4.343   & 0.675 & 0.265 & 0.585 & 0.350 \\
                     & $n=10$ & 26.720 & 17.775 & 13.588 & 7.228  & 0.610 & 0.465 & 0.420 & 0.240 \\
                     & $n=11$ & 64.420 & 47.410 & 22.508 & 11.563 & 0.585 & 0.370 & 0.295 & 0.235 \\
\hline
\end{tabular}
\vspace{-12pt}
\end{table}

\begin{figure}[!ht]
\centering
\subfigure[Uniform instances of size 9]{
\includegraphics[width=0.45\textwidth]{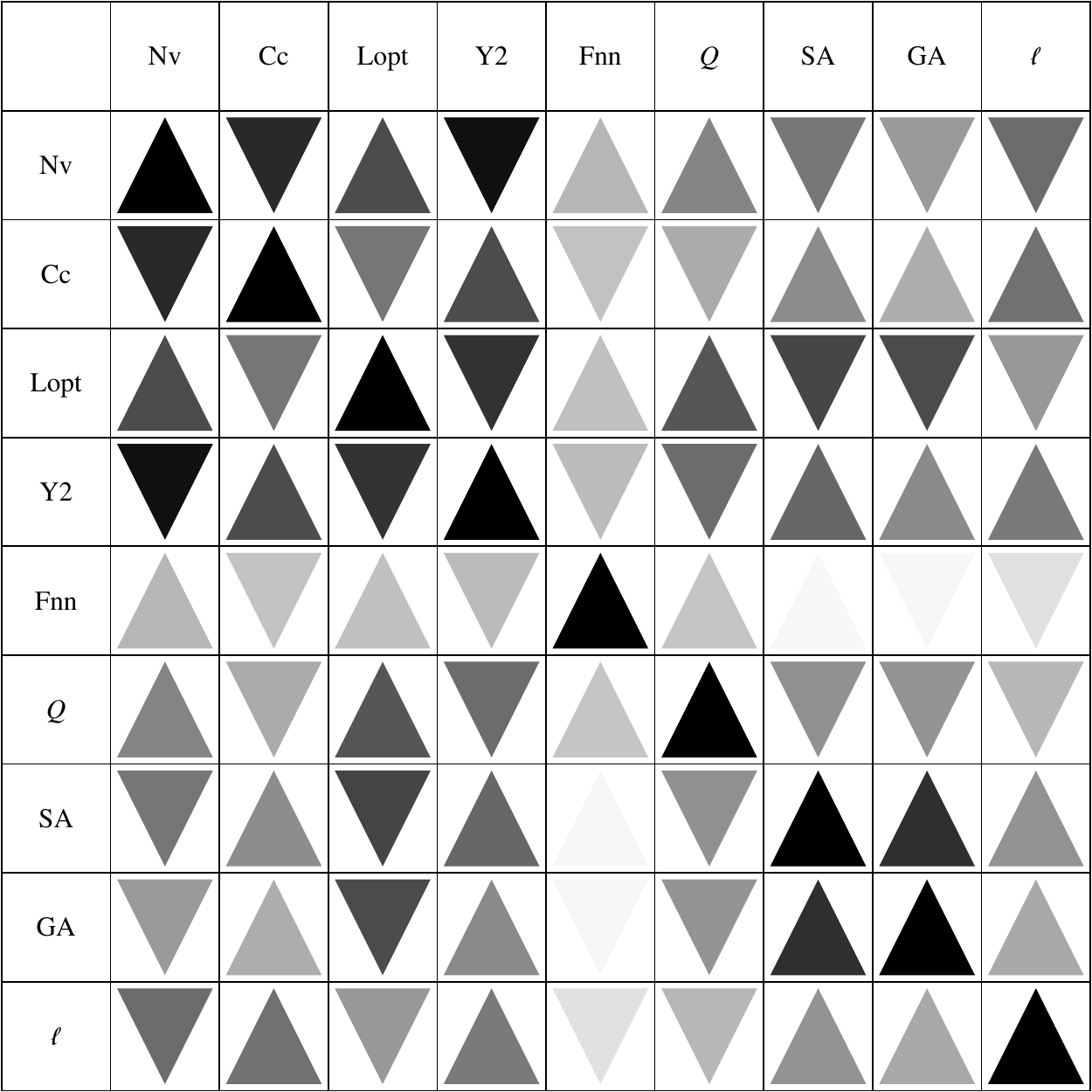}
\label{fig:corr9uni}
}
\subfigure[Real-like instances of size 9]{
\includegraphics[width=0.45\textwidth]{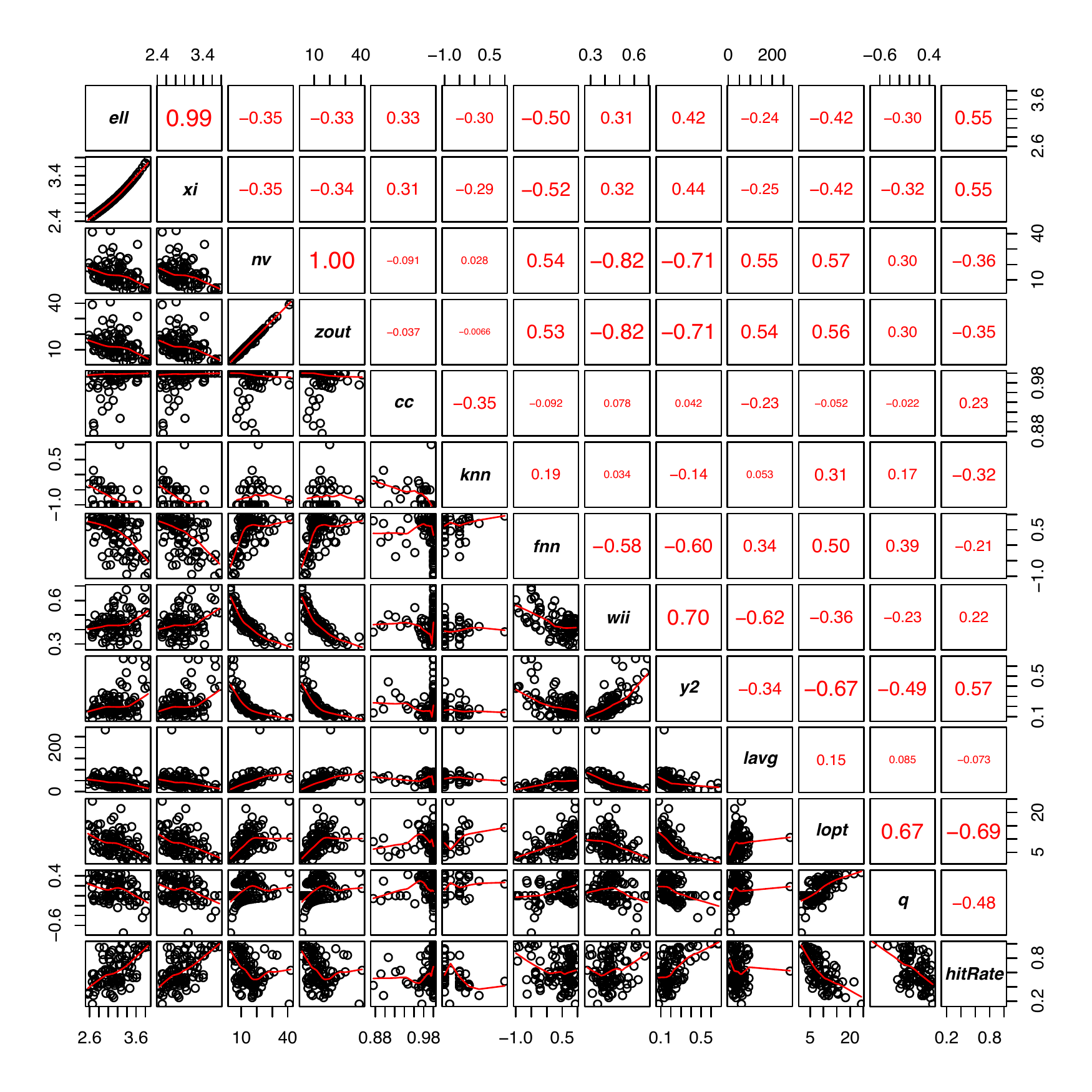}
\label{fig:corr11rl}
}

\subfigure[Uniform instances of size 11]{
\includegraphics[width=0.45\textwidth]{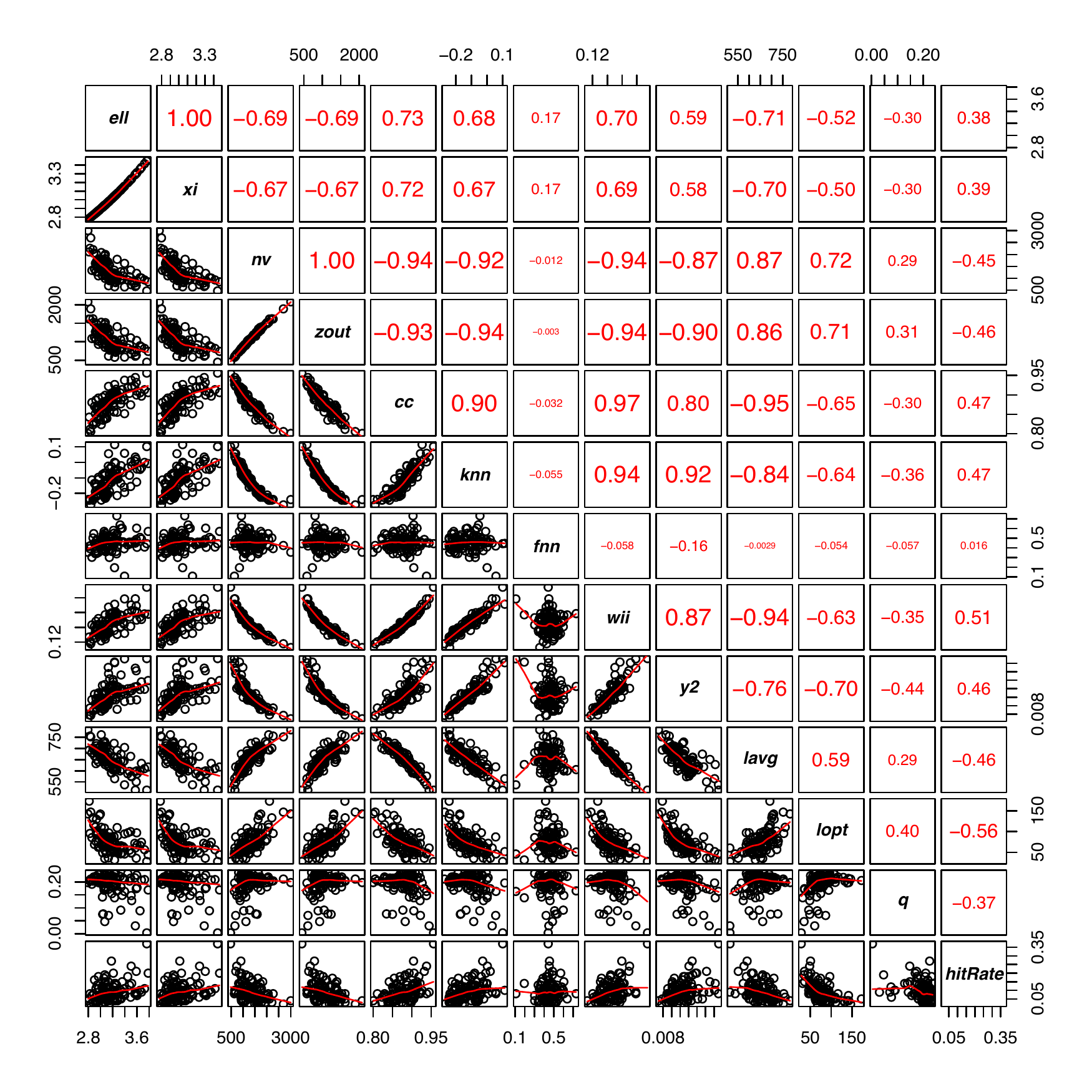}
\label{fig:corr9uni}
}
\subfigure[Real-like instances of size 11]{
\includegraphics[width=0.45\textwidth]{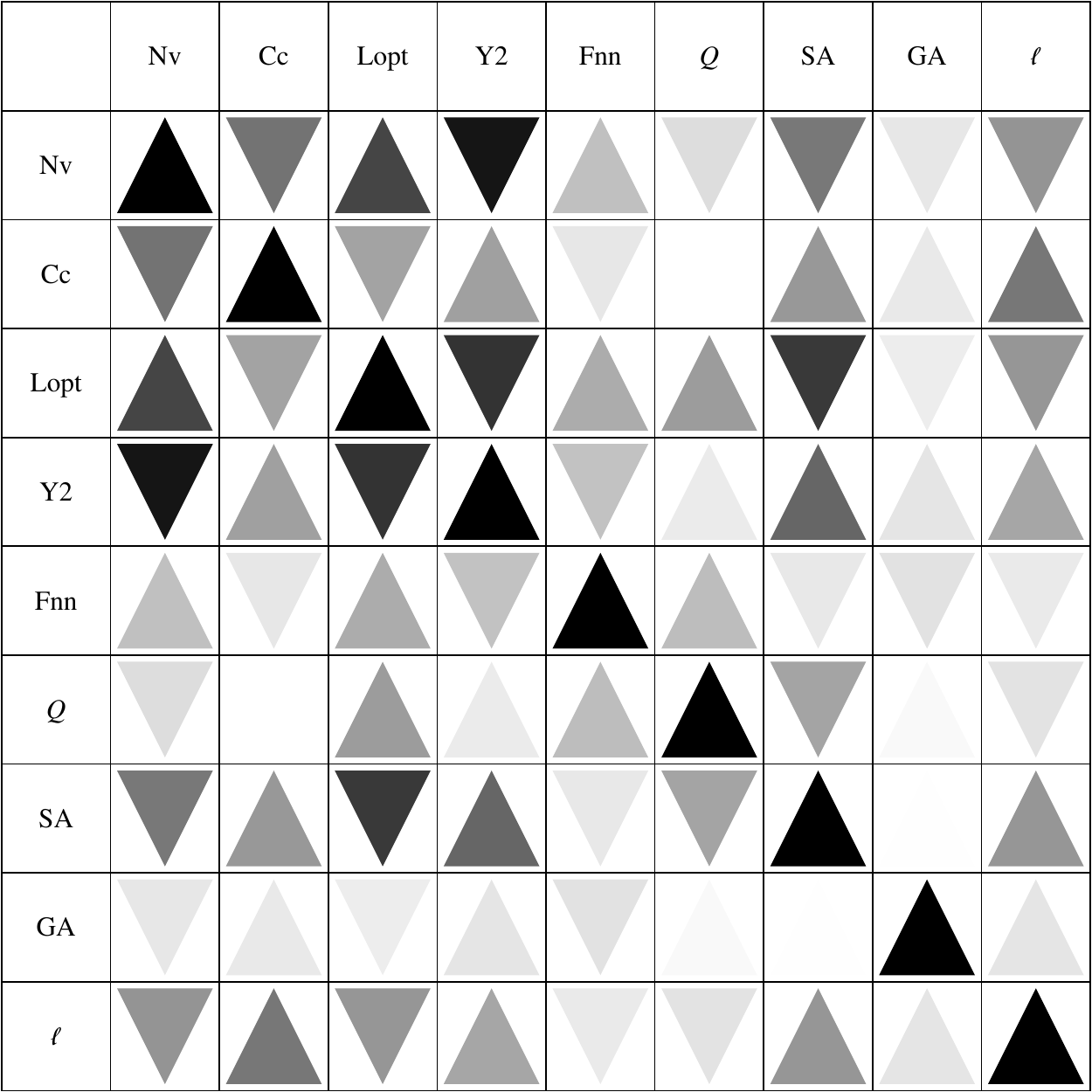}
\label{fig:corr11rl}
}
\caption{Correlations between the measures. An arrow pointing up means positive correlation whereas an arrow pointing down means negative correlation. The absolute value of the correlation is shown in grey scale (the darker the higher).}
\label{fig:correlations}
\vspace{-12pt}
\end{figure}

Let us consider the correlations between the network metrics, the autocorrelation length and the algorithms' performance. These  are shown  qualitatively  in Figure~\ref{fig:correlations} for the instance sizes 9 and  11. The figure shows that  the GA is not correlated to any measure in the real-like instances of large size. Only for the uniform instances and the ones of size 9 there are some significant correlations. We can thus,  conjecture that the measures used in this study are not useful to predict the performance of the GA. A possible explanation is  the presence of the crossover operator, which introduces an additional neighborhood  not used for generating the LONs. On the contrary, the SA algorithm only uses a single move operator (pairwise exchange or swap) which is the same used to generate the LONs. In this case, the Figure reveals correlation with some metrics. In particular, the correlation between the performance of SA and $L_{opt}$ is the highest, which suggests that $L_{opt}$ is the measure that better predicts the behavior of SA.

\begin{figure}[!ht]
\vspace{-12pt}
\centering
\subfigure[SA and $N_v$]{
\includegraphics[width=0.31\textwidth]{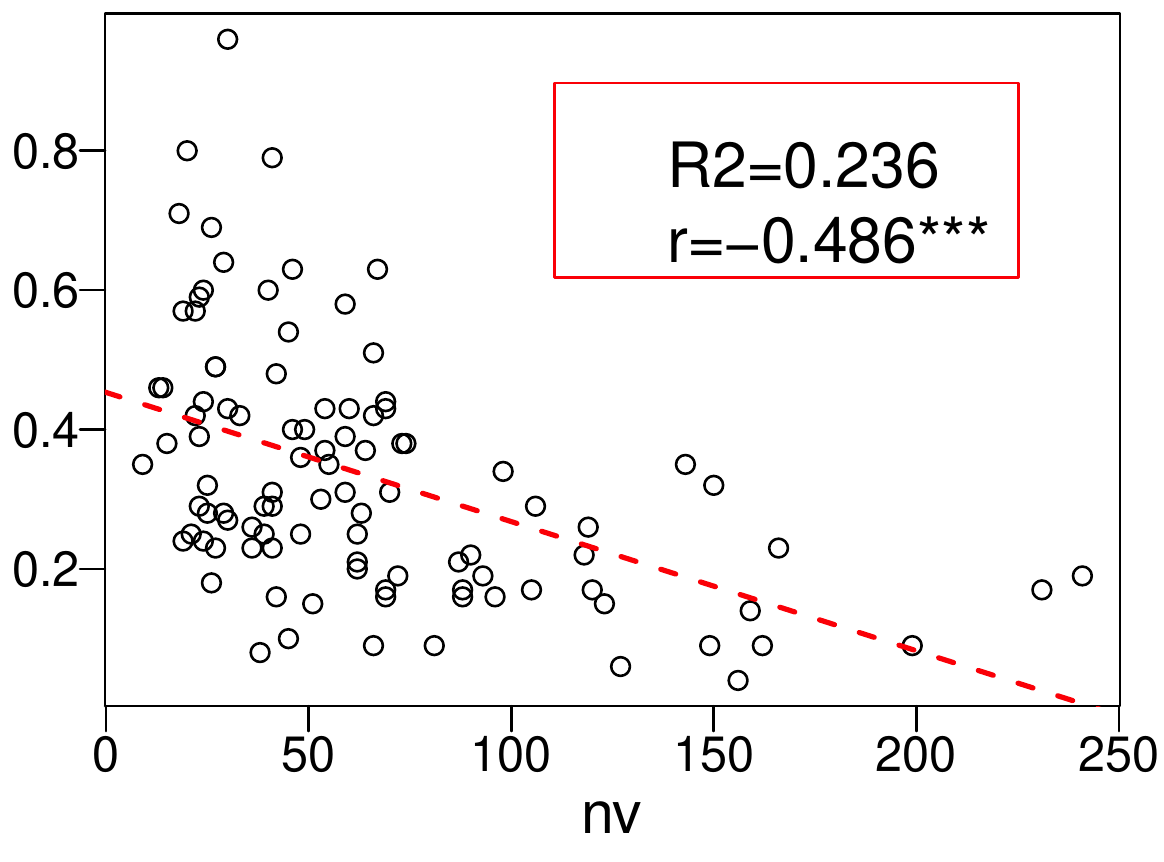}
\label{fig:sa11rl-nv}
}
\subfigure[SA and $Cc$]{
\includegraphics[width=0.31\textwidth]{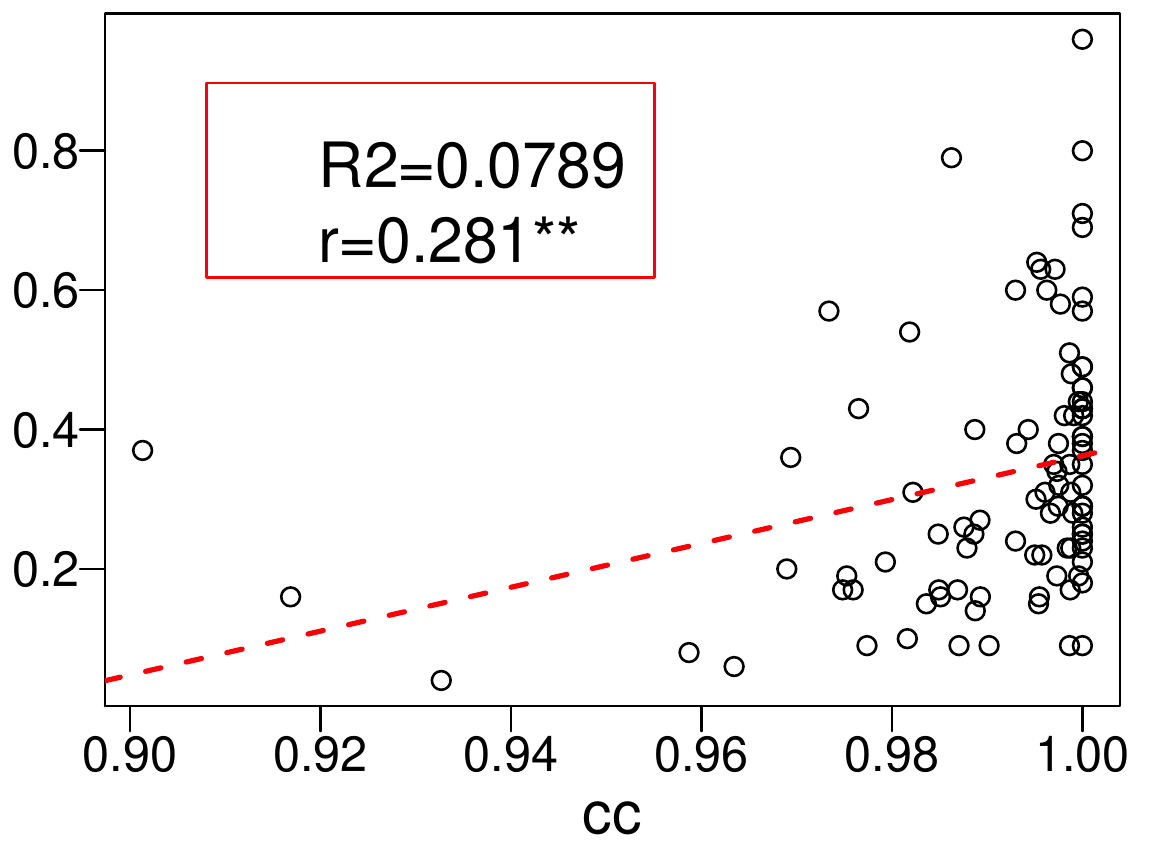}
\label{fig:sa11rl-cc}
}
\subfigure[SA and $L_{opt}$]{
\includegraphics[width=0.31\textwidth]{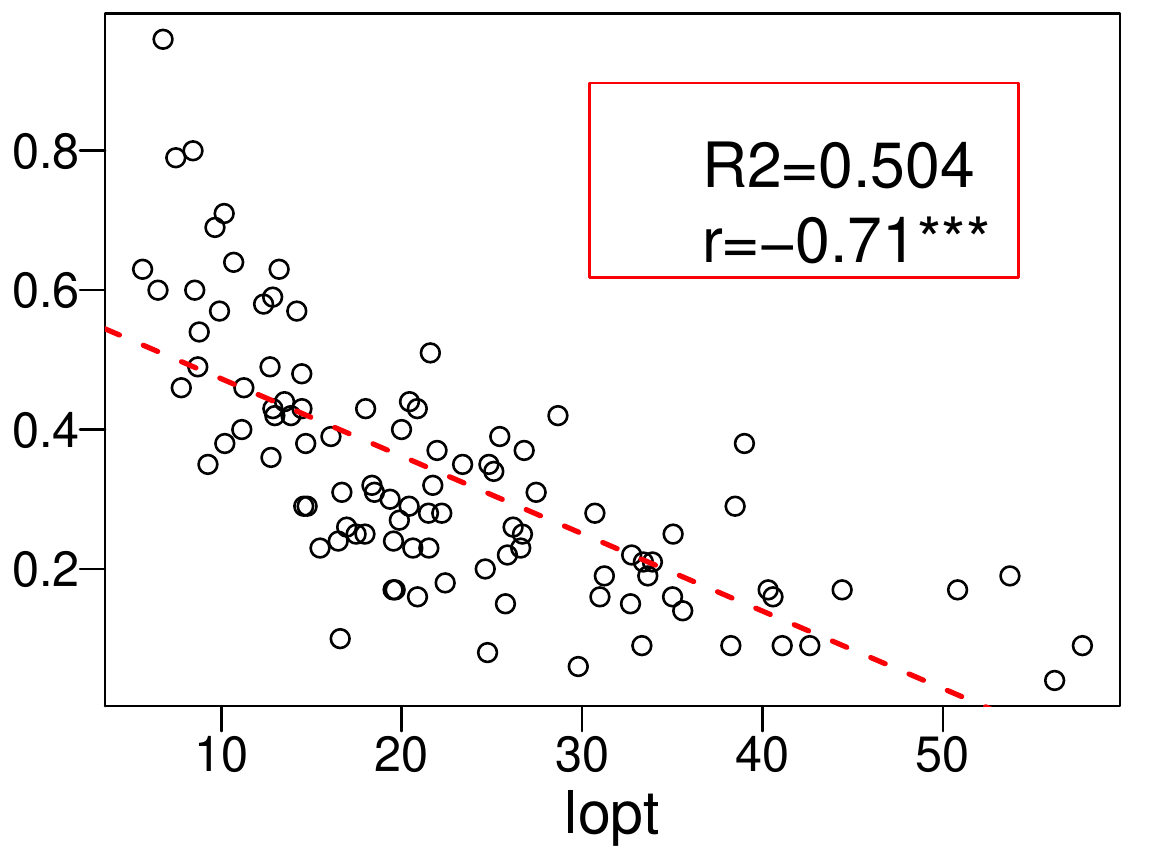}
\label{fig:sa11rl-lopt}
}
\subfigure[GA and $N_v$]{
\includegraphics[width=0.31\textwidth]{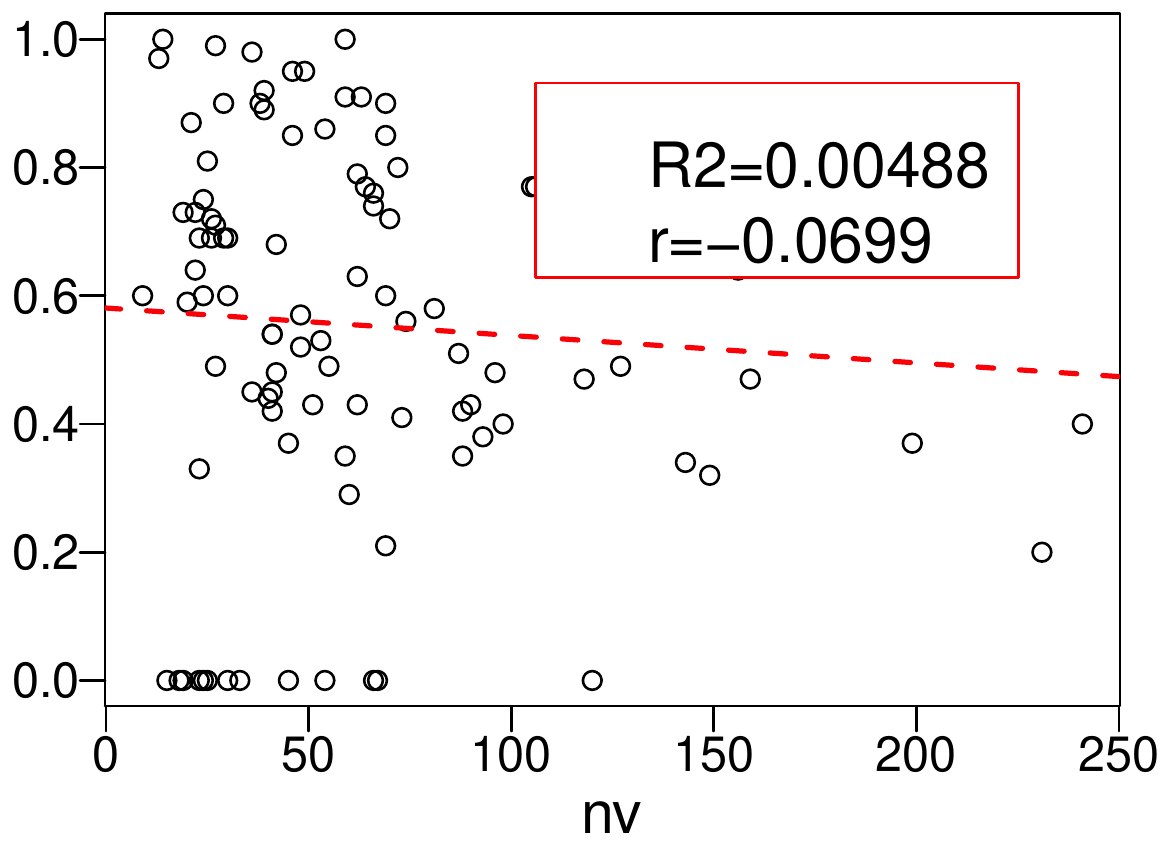}
\label{fig:ga11rl-nv}
}
\subfigure[GA and $Cc$]{
\includegraphics[width=0.31\textwidth]{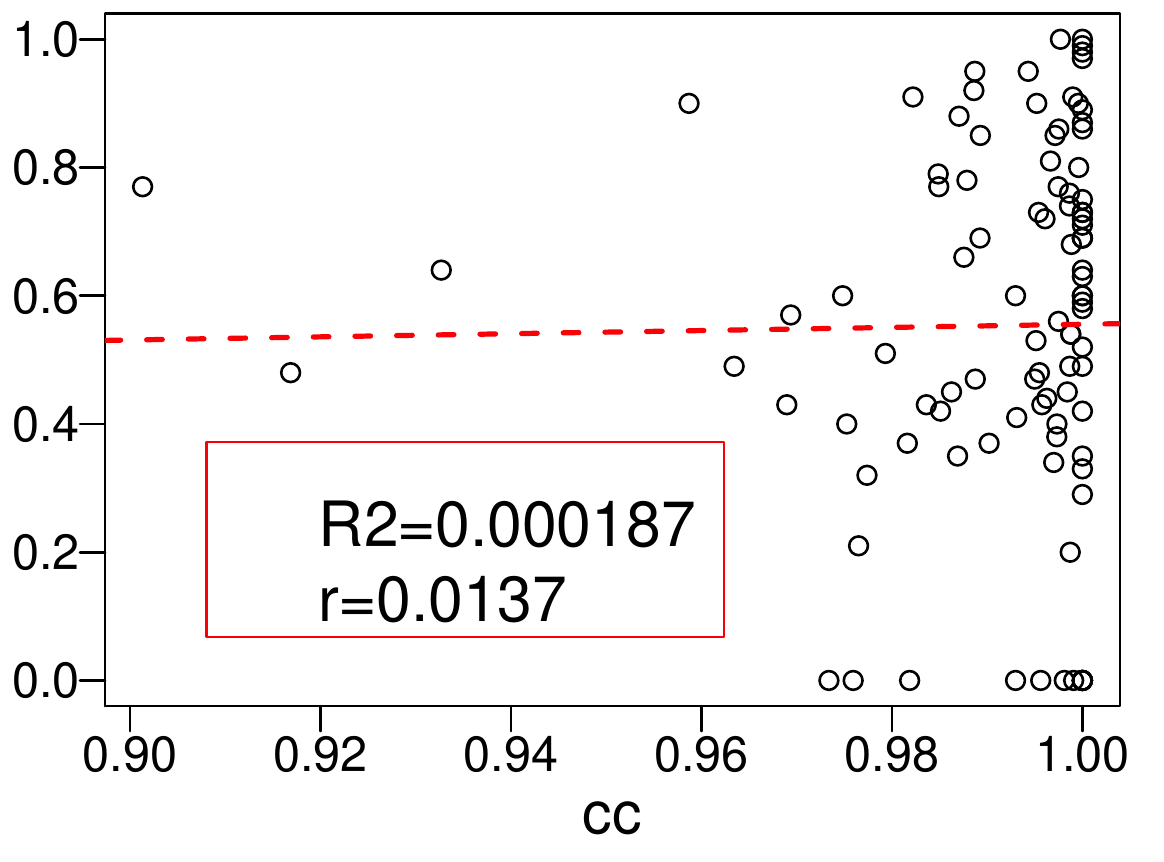}
\label{fig:ga11rl-cc}
}
\subfigure[GA and $L_{opt}$]{
\includegraphics[width=0.31\textwidth]{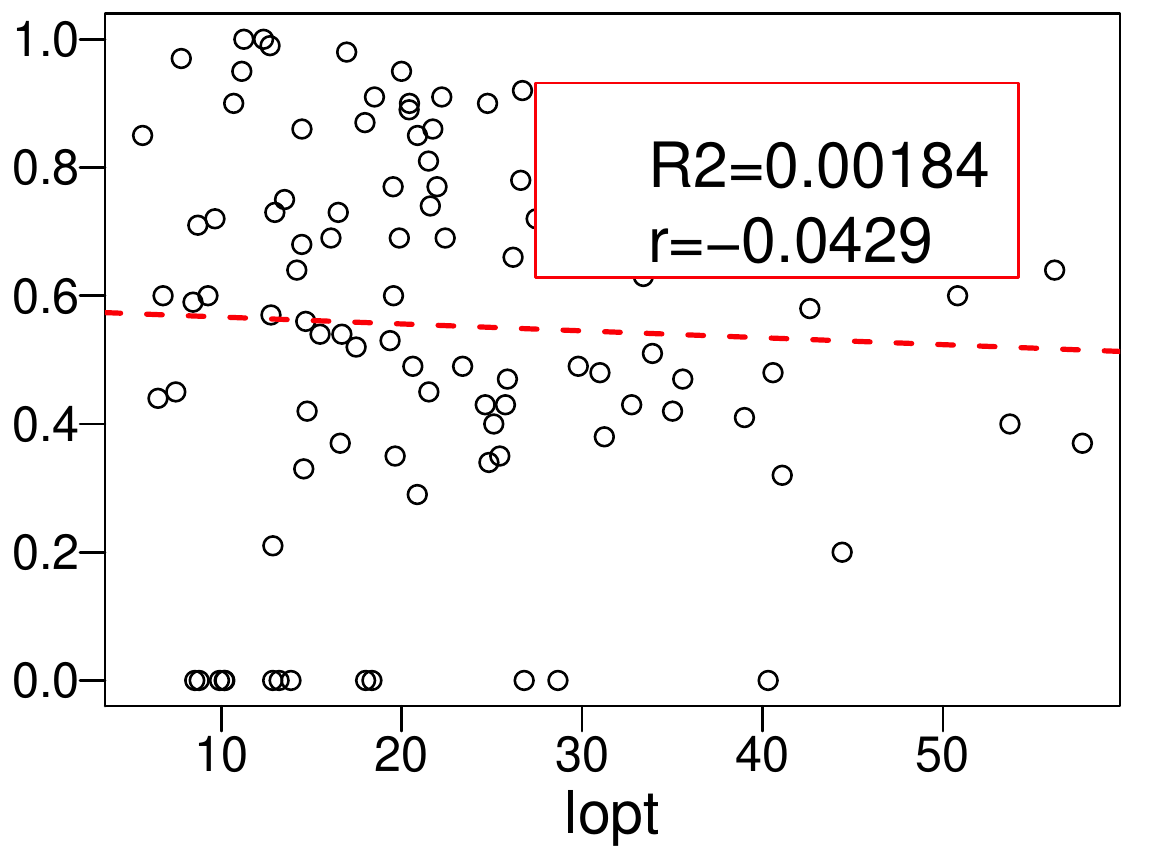}
\label{fig:ga11rl-lopt}
}
\subfigure[SA and $Y_2$]{
\includegraphics[width=0.31\textwidth]{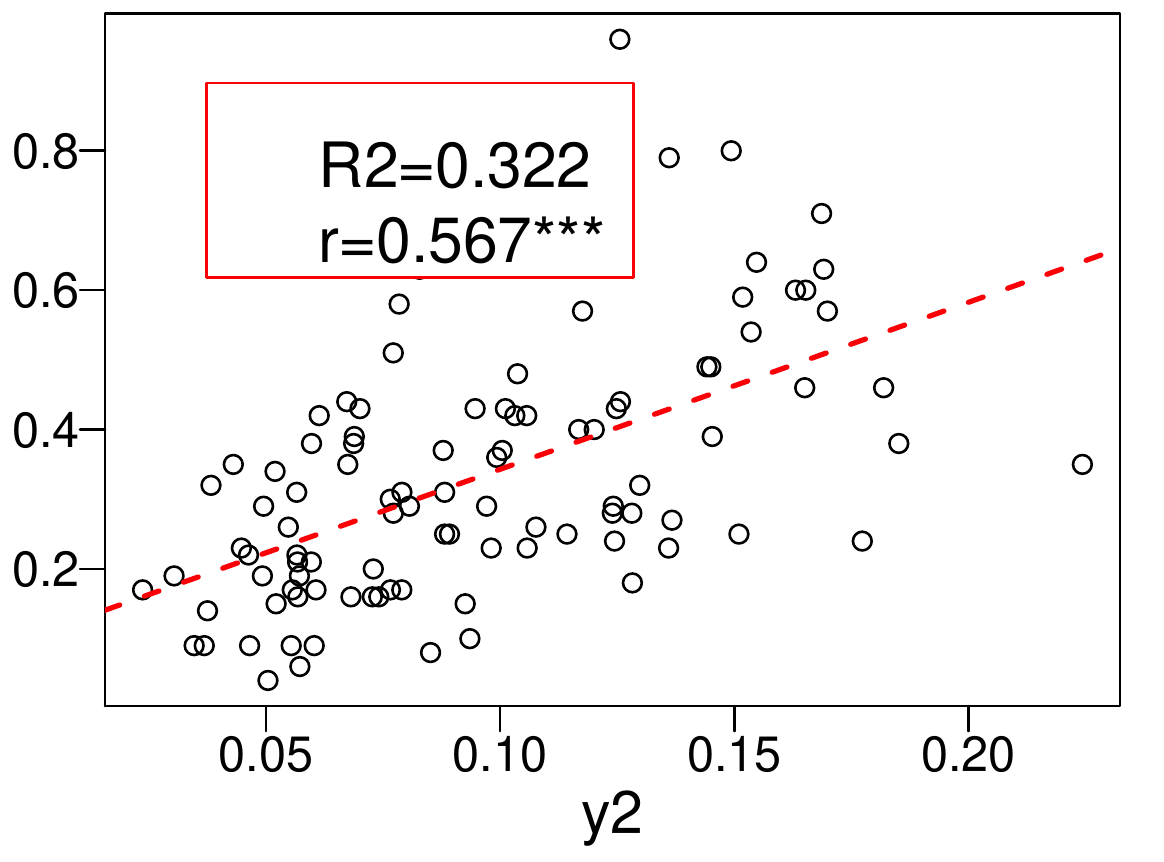}
\label{fig:sa11rl-y2}
}
\subfigure[SA and $Q$]{
\includegraphics[width=0.31\textwidth]{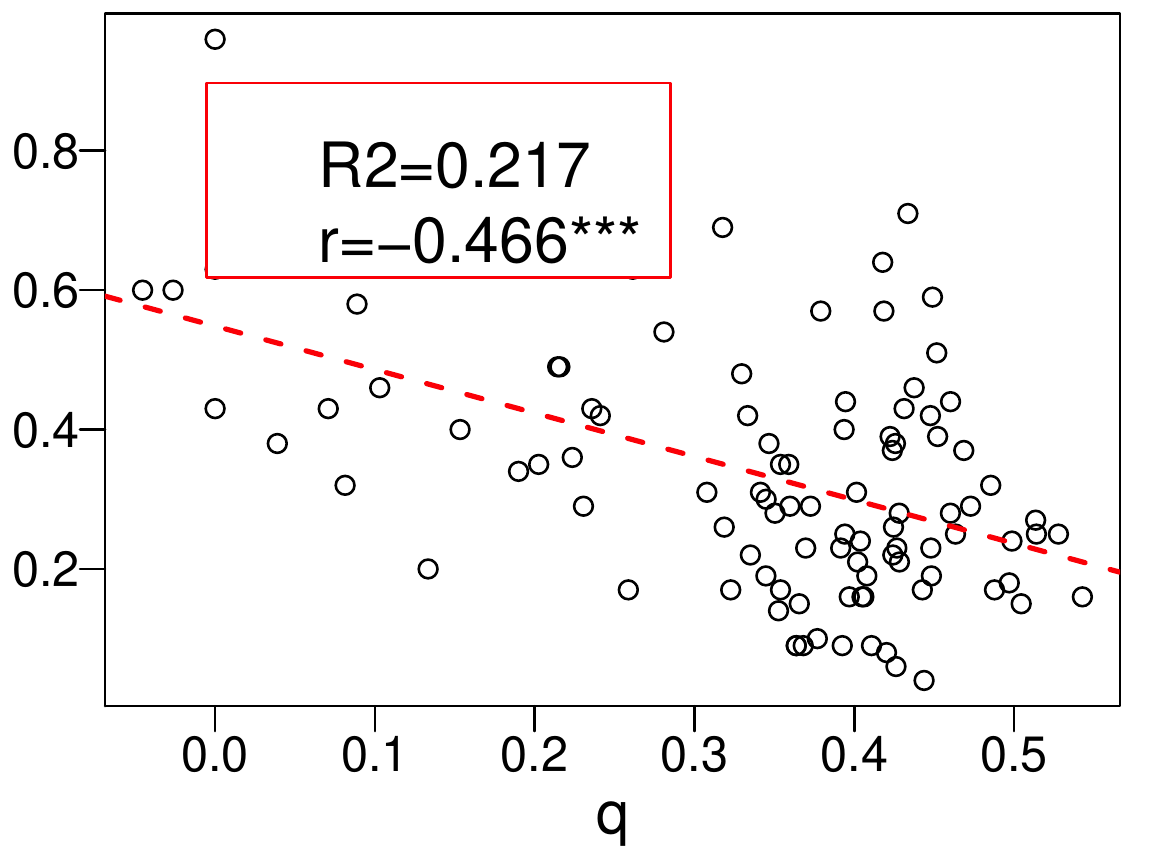}
\label{fig:sa11rl-q}
}
\subfigure[SA and $\ell$]{
\includegraphics[width=0.31\textwidth]{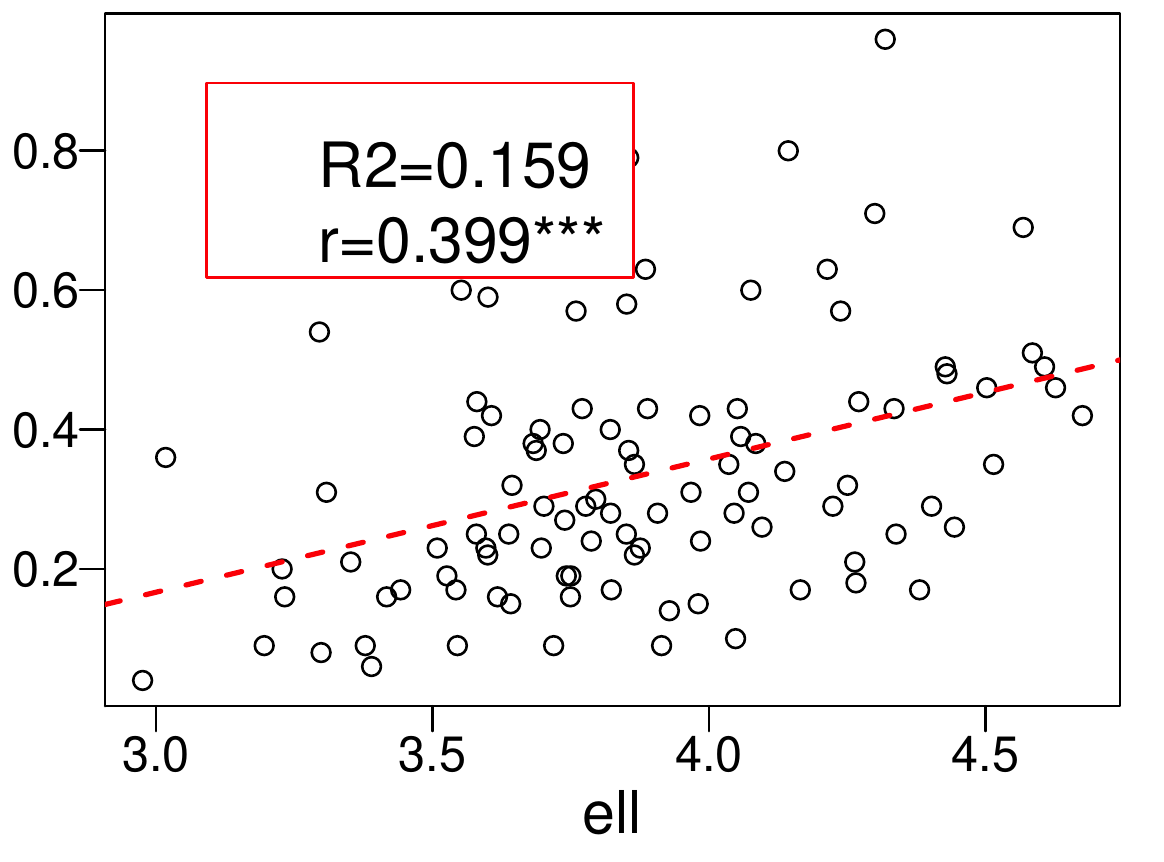}
\label{fig:sa11rl-ell}
}

\subfigure[GA and $Y_2$]{
\includegraphics[width=0.31\textwidth]{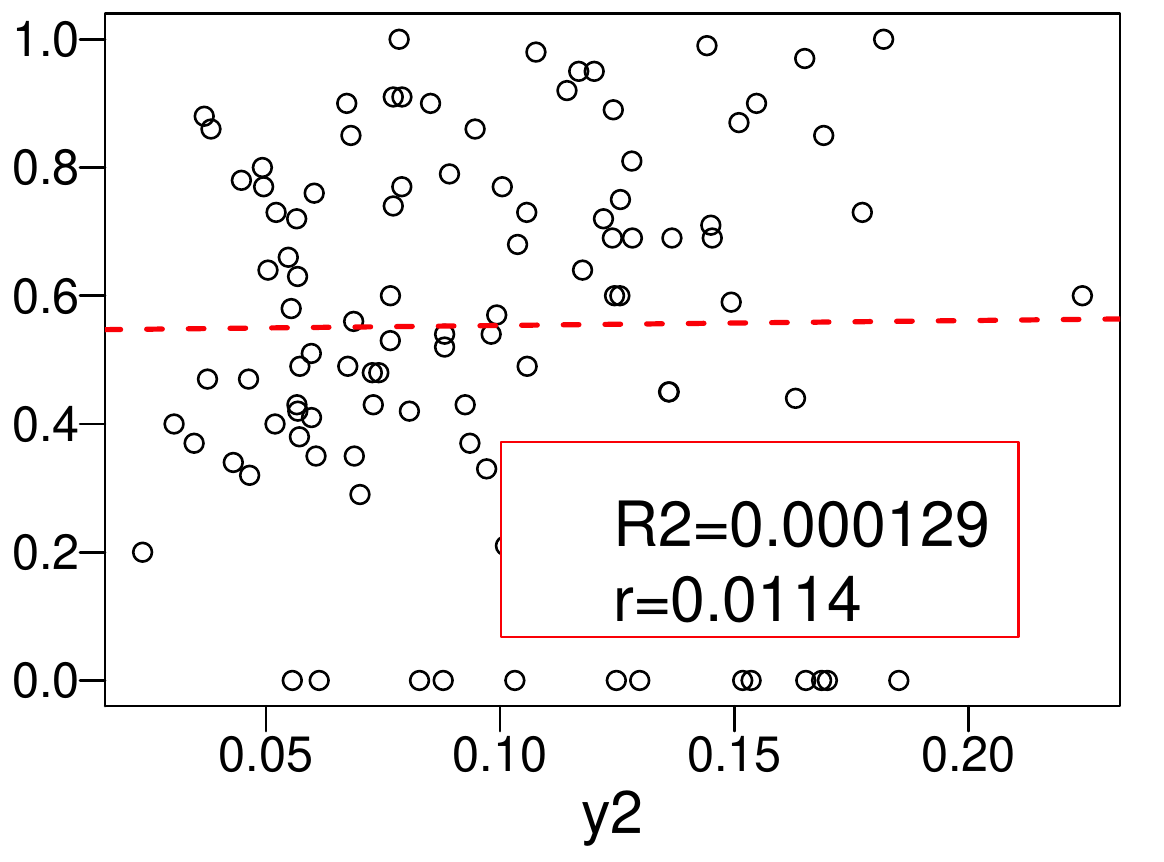}
\label{fig:ga11rl-y2}
}
\subfigure[GA and $Q$]{
\includegraphics[width=0.31\textwidth]{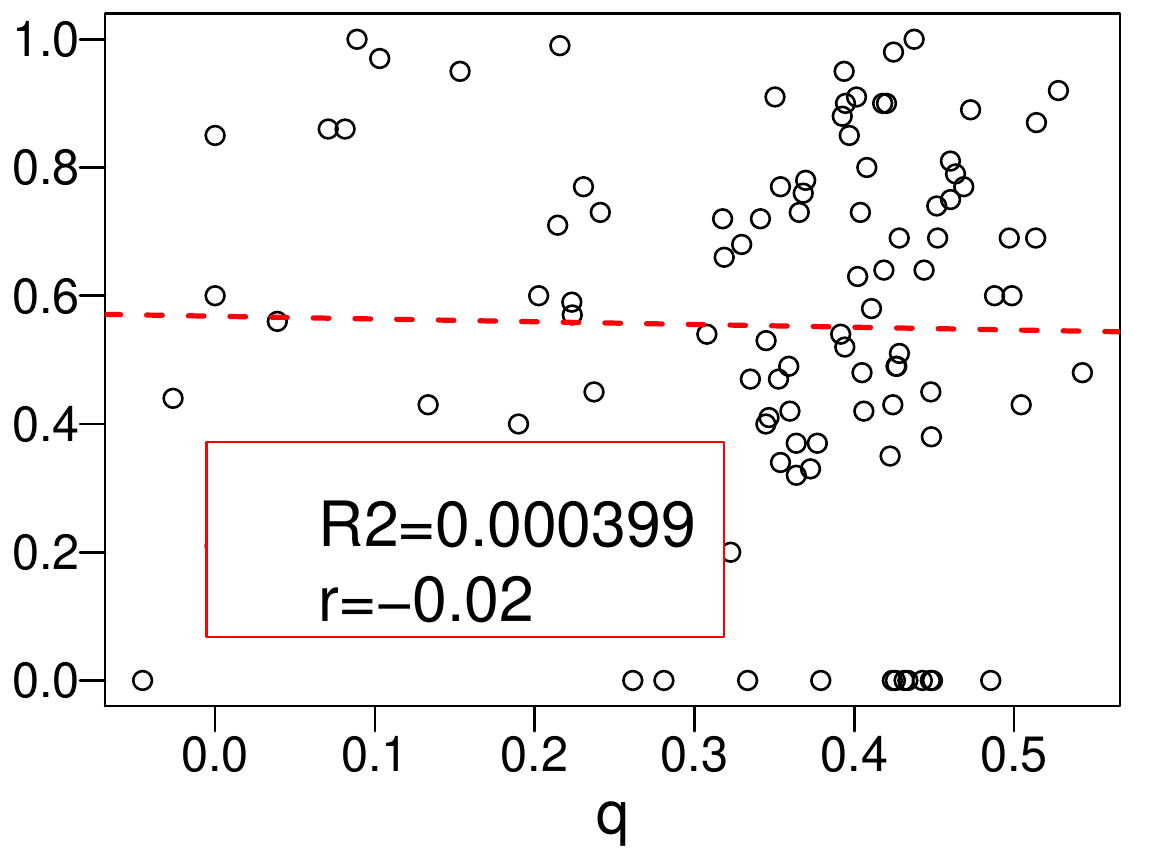}
\label{fig:ga11rl-q}
}
\subfigure[GA and $\ell$]{
\includegraphics[width=0.31\textwidth]{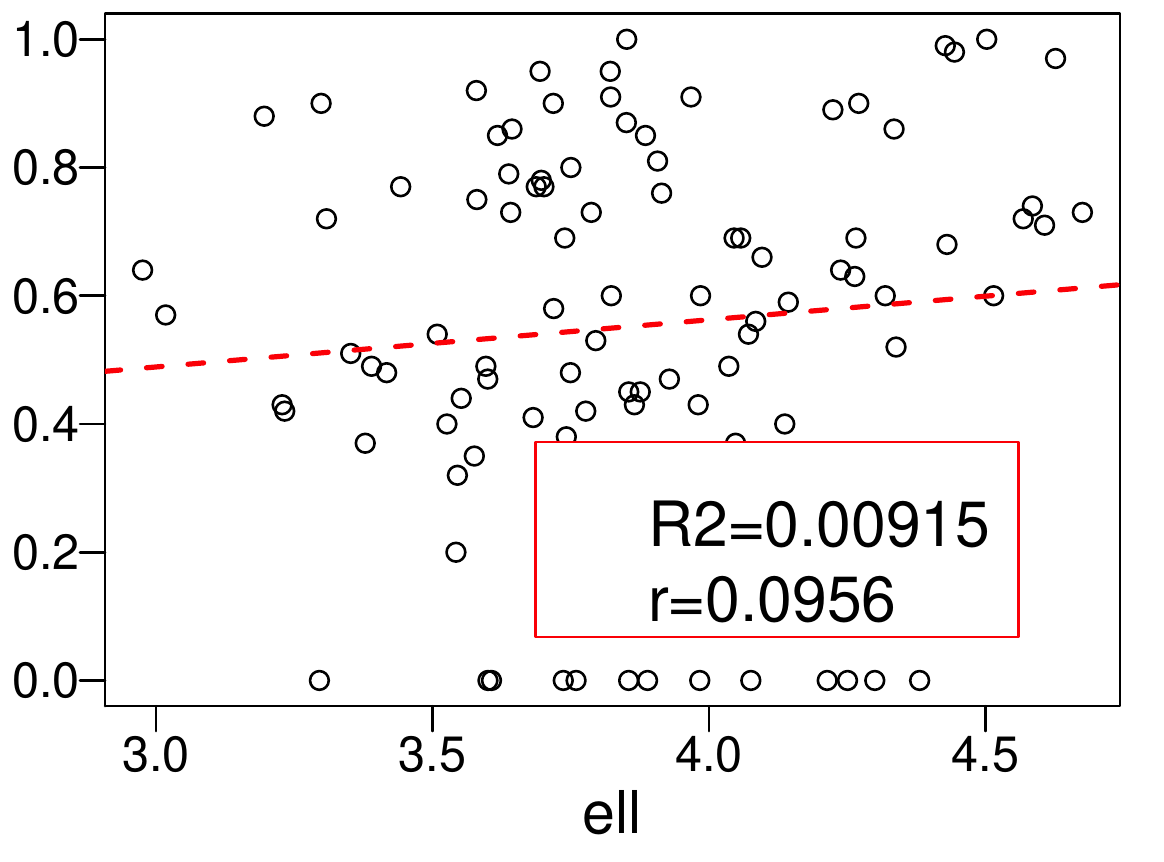}
\label{fig:ga11rl-ell}
}
\caption{Regression analysis for the hit rate of SA and GA against some selected measures for the real-like 11 instances.}
\label{fig:regressions-comp}
\vspace{-20pt}
\end{figure}

In Figure~\ref{fig:regressions-comp} we plot the hit rate against some selected measures for SA and GA in the real-like instances of size 11. The plots of the SA and GA are interleaved in order to compare the results of the regression analysis (the regression line is superimposed on the plot). We can observe how the line has a smaller slope in the case of the GA for all the plots, what explains the low correlation for the GA hit rate and the measures. But we can also observe how $L_{opt}$ is a good predictor of the SA performance.

An interesting observation that contributes to explain the robustness of the GA over the problem size is that while  hit rate for the SA is correlated with the number of local optima, this is not the case for the GA. This suggests that the global search characteristic of a population in GA makes it more robust to the presence of larger number of local optima.

On the uniform  instances, we can observe a positive correlation between the performances of GA and SA, which suggests that the search difficulty is similar for  both algorithms in this case. This is observed for all instance sizes although the  correlation decreases as the size increases. On the real-like instances, the observation is different. In particular,  for the real-like instance of size 11, there is no correlation whatsoever between the performance of both algorithms, which suggests that the search difficulty depends on the algorithm for these instances. In other words, the hard instances for the GA are not the same as the hard instances for the SA and vice versa. We may also speculate that the GA is more efficient at exploiting the more modular structure of the real-like instance, which makes its search dynamic different than that of the SA on these instances.

Regarding the rest of the measures, in general, the correlations are higher in the uniform instances than in the real-like instances. $F_{nn}$  and $Q$  seem to be the less correlated LON measures (this is particularly true for the real-like  instances). The higher correlation coefficients appear between the clustering coefficient, the disparity, the number of nodes and the path to the global optimum. The autocorrelation length seems to be correlated with these measures and with the performance of SA. This is specially interesting, since the autocorrelation length can be computed from the instance data, without the need to exhaustively generate the complete search space (like it happens with the LON measures). The correlation between $\ell$ and the performance of SA suggests that we can use $\ell$ as a measure of problem difficulty, when a trajectory-based search algorithm is used. This idea is also supported by the results of Angel and Zissimopoulos~\cite{Angel:Zissimopoulos2000a}, which provided a positive correlation between an autocorrelation measure and the performance of a SA. The correlation between $\ell$ and the performance of the GA is much smaller, which again indicates that it is harder to predict the performance of the GA using landscape metrics.


Finally, the correlation analysis also provides evidence of the autocorrelation length conjecture. This conjecture claims that the number of local optima is inversely correlated to the autocorrelation length $\ell$~\cite{Stadler2002}. In~\cite{Chicano2012aml} some results were presented that supported the autocorrelation length conjecture. In that work, the correlation between the number of local optima and $\ell$ was between $-0.1640$ and $-0.3256$. In our case the correlation is higher (in absolute value), in the range from $-0.3729$ to $-0.7187$. This support of the conjecture is higher in the uniform instances than in the real-like instances of the same size.

\vspace{-12pt}
\section{Discussion}
\label{discussion}
\vspace{-12pt}

We conducted a large statistical correlation study considering QAP instances of different types and sizes, a number of landscape metrics and the performance of two widely known search heuristics. Our study also brings together two recent developments in combinatorial landscape analysis, with the aim of shedding new light on  the relationships between the landscape structure and the performance of heuristic search algorithms. Our study confirms that the real-like instances are easier to solve by heuristic search algorithms. Clearly, in these problems, the number of local optima in the landscape is a much better predictor of search difficulty than the size of the search space. As we already knew from the study in \cite{QAPcec10},  for a fixed problem dimension, real-like instances  result in much smaller networks (small number of vertices); the size difference between the two types of QAP  instances grows almost exponentially with the problem dimension.

Overall, the GA was a stronger algorithm to solve all the studied classes of QAP instances. Moreover, the GA is more robust to the increase in problem size. Interestingly, the performance of SA and GA is correlated for the uniform instances, but this is not the case for the real-like instances. Which suggests that the GA is better at exploiting the more clustered structure of the real-like instances.  However, predicting the performance of the GA seems to be a harder task than predicting the performance of SA. GAs are more complex algorithms as they incorporate a population and a recombination operator. In particular, we found some network metrics such as the average distance to the global optima and the number of local optima, which are good predictors of the SA performance, but less so for the GA. The question is still open for a better understanding and prediction of the GA performance. 

Finally, our study provides supporting evidence of the correlation length conjecture indicating that the number of local optima is inversely correlated to the correlation length. This is an interesting contribution, as using elementary landscape decomposition the autocorrelation length for QAP instances can be exactly calculated from the instance data~\cite{Chicano2012aml}.

More detailed studies, additional metrics, sampling approaches to extract the LONs and larger landscapes are required to better understand and predict search difficulty in combinatorial optimization. Our study, however, is a first step that incorporates new landscape metrics coming from the field of complex networks, and try to correlate them with both previously studied  landscape metrics and the performance of heuristic search methods.

\vspace{-12pt}
\section*{Acknowledgements}
\vspace{-12pt}

This work was partially funded by the Spanish Ministry of Science and Innovation and FEDER under contract TIN2011-28194 (the roadME project) and the Andalusian Government under contract P07-TIC-03044 (DIRICOM project).

\vspace{-12pt}

\bibliographystyle{plain}

\end{document}